\documentclass[11pt]{article}
\usepackage{acl}

\usepackage{times}
\usepackage{latexsym}
\usepackage[T1]{fontenc}
\usepackage[utf8]{inputenc}
\usepackage{microtype}
\usepackage{inconsolata}

\usepackage{graphicx}
\usepackage{amsmath}
\usepackage{amssymb}
\usepackage{booktabs}
\usepackage{array}   
\usepackage{tabularx} 
\usepackage[x11names]{xcolor} 
\graphicspath{{figures/}{figures/E3/}{figures/E8/}{figures/E10/}}

\newcommand{\AlgOne}{\textsc{Mean-SVD}}
\newcommand{\DiffPCA}{\textsc{DiffPCA}}
\newcommand{\LEACE}{\textsc{Leace}}
\newcommand{\Wiener}{\textsc{Wiener}}

\newcommand{\MGSM}{\textsc{MGSM}}

\newcommand{\MATHD}{\textsc{MATH}}           
\newcommand{\Pooled}{\textsc{Activation-PCA}} 
\newcommand{\Rand}{\textsc{Random}}          

\title{Beyond Accuracy: What Language-Subspace Interventions Capture, Remove, and Improve}

\author{
  Shaanil Punglia\textsuperscript{1},
  Dadhichi Telwadkar\textsuperscript{1},
  Vaibhav Rathore\textsuperscript{2,1}, \\
  \textbf{Siddharth Sureka\textsuperscript{1},
  Neha Bhargava\textsuperscript{1}} \\
  \textsuperscript{1}Motilal Oswal Financial Services, India \\
  \textsuperscript{2}Indian Institute of Technology Bombay, India \\
}

\begin{document}
\maketitle

\begin{abstract}
Language-subspace interventions aim to improve multilingual reasoning by suppressing language-related components in mid-layer activations. Using \MATHD{} and \MGSM{} across four open-weight LLMs, we ask three questions: do estimated language subspaces capture translation-associated variation? Does projecting them out remove language information? And do they improve accuracy without compromising response language? For the first question, we build subspaces using the existing \AlgOne{} construction and our \DiffPCA{}. Both capture substantial translation-associated variation and show similar patterns across model depth. For the second, we project out each subspace and train linear probes to classify language from the projected activations. The probes still identify language almost perfectly, showing that neither subspace exhaustively captures language information. For the third, we evaluate several intervention configurations on Qwen2.5-7B, by varying subspace creation methods and intervention operators. We find that the intervention that most improves \MGSM{} accuracy sacrifices response-language fidelity, with only one intervention improving joint accuracy--answers that are correct and in the target language--slightly, by 1.60 points. Interestingly, subtle changes in the prompt alone can reproduce the accuracy gains albeit with significant damage to language fidelity. We argue that causal interventions must be judged on more than accuracy, to fully account for language substitution trade-offs.

\end{abstract}

\section{Introduction}
Mechanistic interventions can produce the desired metric improvement for an unintended reason. In multilingual reasoning, an intervention can
raise accuracy simply by moving generation into English or Chinese. The answer may be mathematically
correct, yet the model has silently changed the task. Multilingual reasoning therefore offers a
useful stress test for causal interpretability: a behavioral gain need not validate the mechanism
an intervention was designed to target.

Language-subspace ablation makes the distinction concrete. \citet{zhao2025less} estimate a low-rank
span from multilingual activation means, suppress it in middle layers, and amplify it near the
output. This raises three separate questions: Does the span track language variation? Does removing the span eliminate linearly decodable language information? Does editing it improve the requested behavior? These are questions of
geometry, containment, and behavioral effect, and each requires its own evidence.

We test those properties separately. We compare mean-based and translation-difference subspaces,
probe what projection disrupts and what language information survives, and evaluate generation with
\emph{joint accuracy}: an answer must be correct and remain in the problem language. Additionally,
we construct prompt-language controls that leave the problem untouched but switch its worked
examples and answer cue to English or Chinese, testing whether contextual language alone can
reproduce the apparent gain.

The evidence reveals a consistent split. The subspaces summarize translation structure without
containing language exhaustively, and the accuracy-maximizing edit performs worse once output language is part of
the task. Prompt-language switches reproduce the trade-off, suggesting that the gain comes from
changing the elicited linguistic regime rather than from activation editing itself. The broader
lesson is methodological: a clean intervention in activation space is not yet a clean causal explanation. A strong causal account requires evidence connecting the edited geometry, the information affected, and the resulting behavioral change.
\section{Related work}
\label{sec:related}

Language-subspace interventions grow from evidence that multilingual representations contain both
shared and language-sensitive geometry. Low-rank analyses recover language-associated structure
\citep{xie2022discovering,chang2022geometry}, while neuron-level studies identify units that predict
or steer output language \citep{tang2024neurons,zhao2024multilingualism}. Intermediate states can
also favor English concept realizations even when the input is not English
\citep{wendler2024llamas}. Activation-patching results separately suggest that concept
representations can be partially decoupled from output language \citep{dumas2025tongue}.
Together, these results motivate compact language directions, but do not show that such directions
contain language exhaustively or control how they are used.

Activation steering turned this descriptive geometry into an intervention surface: low-dimensional
edits can redirect high-level behavior \citep{turner2023steering,rimsky2024steering}.
\citet{zhao2025less} turn this idea into an intervention: they estimate a language subspace from
translated mathematics problems, suppress it in middle layers, amplify it near the output, and
evaluate its effects across multilingual tasks. Their work is the direct premise of ours, while
leaving open how effects depend on the subspace construction and removal operator.

Removal, however, is not automatically explanation. Probe accuracy can reflect readout capacity
rather than representation quality \citep{hewitt2019probes}, and information accessible to a
readout need not be behaviorally used by the model \citep{elazar2021amnesic}. Iterative Nullspace
Projection (INLP) introduced repeated removal of probe directions to guard representations from
linear prediction \citep{ravfogel2020null}, and amnesic probing connected removal to behavioral
counterfactuals \citep{elazar2021amnesic}. \LEACE{} later provided closed-form linear erasure
\citep{belrose2023leace}; subsequent comparisons stress that a useful causal intervention must
remove the target selectively rather than broadly distort the representation
\citep{dobrzeniecka2025improving}. This lineage motivates our retrained probes and operator
comparison: defeating a fixed probe, exhausting linearly decodable information, and changing
generation are distinct claims.

Behavioral evaluation adds a final complication. In multilingual mathematics, the language of
demonstrations and cues affects accuracy, and determines the language the model is prompted to
reason in \citep{shi2023language}.
Accordingly, an intervention cannot be judged by correctness alone when remaining in the problem
language is part of the task. Prior work motivates each link in this chain, but usually examines
them separately. We compare mean-based and translation-difference subspaces, expand the operator
axis beyond orthogonal projection, probe what survives removal, evaluate complete task success, and
ask whether prompt language reproduces the same pattern without activation editing.

\section{Setup}
\label{sec:setup}

\subsection{Datasets}

We use the same mathematical problems expressed in multiple languages, so that problem content
stays fixed while language changes. The \MATHD{} dataset contains 12,500 competition-level
problems \citep{hendrycks2021math}. Following \citet{zhao2025less}, we use its 7,500-problem
training split and aligned machine translations from English into ten additional languages to
estimate language subspaces.
\MGSM{} provides 250 grade-school problems professionally translated across the same eleven languages,
with native-language worked examples for generation \citep{shi2023language,cobbe2021gsm8k}.
We use it to test the subspaces on a different corpus and to measure their behavioral effects.
Thus \MATHD{} supplies scale for finding directions, while \MGSM{} asks whether those directions
transfer and matter during generation.

We use four open-weight models of similar scale to carry out our study: Qwen2.5-7B \citep{qwen2024qwen25}, Qwen3-8B
\citep{qwen2025qwen3}, R1-Distill-7B \citep{deepseek2025r1distill}, distilled from
DeepSeek-R1 \citep{deepseek2025r1}, and GLM-Z1-9B
\citep{zai2025glmz1}.

\subsection{Language subspaces}

The same problem expressed in two languages lets us ask a clean geometric question: how does its
representation move when the linguistic realization changes? Let $h_{i,k}^\ell\in\mathbb{R}^d$
be the residual-stream activation at the final prompt token for problem $i$, language $\ell$, and
layer $k$. We fit a rank-$r$ basis $U$ independently at each layer and omit the layer index from
the notation below.

\paragraph{Translation differences.}
For two renderings of the same problem, define
\begin{equation}
  v_i^{s\rightarrow t}=h_i^t-h_i^s.
  \label{eq:translation}
\end{equation}
These vectors hold problem identity fixed. They are therefore a direct test of whether a proposed
subspace follows translation. Because translation also changes tokenization, wording, and possibly
the representation of the mathematics, we call this \emph{language-associated variation}, not
pure language.

\begin{figure*}[t]
  \centering
  \includegraphics[width=0.9\textwidth]{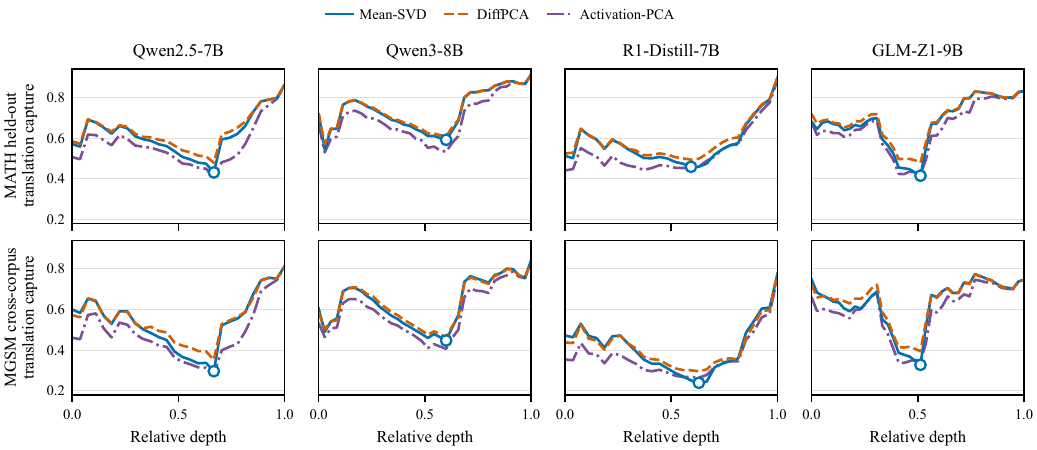}
  \caption{Rank-10 capture of translation differences over model depth. Columns are models and
  rows are held-out \MATHD{} and cross-corpus \MGSM{}. Circles mark the \AlgOne{} interior
  minima.}
  \label{fig:rho}
\end{figure*}

\paragraph{Mean-based subspace (\AlgOne{}).}
\AlgOne{} compresses each language to one mean activation, then factorizes the resulting matrix
into a shared anchor and leading directions that distinguish the language means
\citep{piratla2020efficient,xie2022discovering,zhao2025less}. Averaging suppresses individual
problems and emphasizes stable offsets between languages. With eleven language means, its
language-varying component has rank at most ten.

\paragraph{Within-problem subspace (\DiffPCA{}).}
\DiffPCA{} instead keeps problem identity explicit. For each problem, it subtracts the mean of its
translations and normalizes the remaining language-specific residuals:
\begin{equation}
  R_i^\ell=h_i^\ell-\frac{1}{L}\sum_{t=1}^{L}h_i^t,
  \qquad z_i^\ell=\frac{R_i^\ell}{\max(\|R_i^\ell\|,\epsilon)}.
\end{equation}
Its basis is the top $r$ principal components of the centered $z_i^\ell$. This favors directions
repeatedly traversed by translations rather than differences between languages only in the
aggregate, and it permits ranks above ten. The two estimators therefore encode different ideas:
\AlgOne{} treats language as a stable population offset, whereas \DiffPCA{} treats it as a
recurring within-problem movement. Agreement across them suggests that the observed
translation-associated geometry is not an artifact of either construction.

\paragraph{Controls.}
We compare both with high-variance directions learned without language labels (\Pooled{}) and a
random orthonormal span (\Rand{}). \Pooled{} asks whether the apparent language structure is
specific, or merely a consequence of projecting onto the model's dominant directions. Full algorithms and fitting details appear in
Appendix~\ref{app:construction}.

\section{How much translation variation is captured?}
\label{sec:geometry}

\begin{figure*}[t]
  \centering
  \includegraphics[width=\textwidth]{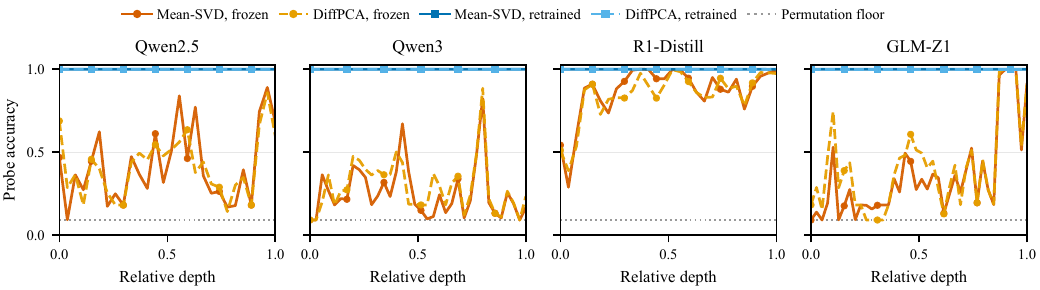}
  \caption{The retrained probe's accuracy curves remain near the ceiling even where the frozen
  probe's accuracy curves approach the permutation floor (gray). Although the gap varies
  irregularly with layer depth, it remains pronounced for all models except R1-Distill, whose
  frozen probe is comparatively robust.}
  \label{fig:probe-depth}
\end{figure*}

For a translation vector $v$ defined in Equation~\ref{eq:translation}, we measure the fraction of
its squared norm captured by a subspace with orthonormal basis $U$:
\begin{equation}
  \rho_U(v)=\frac{\|U^\top v\|^2}{\|v\|^2}.
  \label{eq:capture}
\end{equation}
We average over evaluation problems and all 55 unordered language pairs.
Appendix~\ref{app:geometry} gives the data partitions and full results.

Figure~\ref{fig:rho} shows a decline in rank-10 capture through intermediate layers followed by
an increase toward the output. Both estimators exhibit this pattern across all four models on
held-out \MATHD{} and cross-corpus \MGSM{}. Their interior minima coincide within each
model--corpus pair, at 51--67\% of model depth. On \MGSM{}, \AlgOne{} captures
24.0--44.8\% of translation variation at these minima, and \DiffPCA{} captures 29.6--46.2\%.
Thus, on this corpus, more than half of the variation remains outside either rank-10 span.

\Pooled{} provides a stronger comparison than random directions. At the \MGSM{} minima,
it captures 26.5\% for R1-Distill and 34.6\% for GLM, exceeding \AlgOne{}'s 24.0\% and
32.8\%, respectively. \DiffPCA{} captures more than \Pooled{} at all four models'
\MGSM{} minima, but substantial capture is already obtained without language labels or
within-problem residualization. Translation differences therefore overlap substantially with
high-variance activation directions; capture alone does not establish language specificity.

Increasing \DiffPCA{}'s rank extends coverage beyond the rank-ten limit of \AlgOne{}.
At Qwen2.5 layer 18 on held-out \MATHD{}, capture rises from 47.9\% at rank ten to
74.0\% at rank 100, compared with 2.8\% for a rank-100 random span.
We next test whether language remains linearly decodable after removing these spans.

\section{Language decodability after projection}
\label{sec:probes}

We test whether language identity remains linearly decodable after removing the estimated
subspaces. Using subspaces fitted on \MATHD{}, we ablate zero-shot \MGSM{} activations at
each layer:
\begin{equation}
  h'=h-\lambda UU^\top h,
  \label{eq:ablation}
\end{equation}
where $U$ is an orthonormal subspace basis. At $\lambda=1$, the component within the subspace
is removed completely. For $0\leq\lambda<1$, the transformation is invertible and preserves
linear separability in principle. We evaluate each layer independently, without propagating
edited activations through subsequent blocks.

We train eleven-way logistic-regression classifiers using five-fold cross-validation, keeping
all translations of each problem in the same fold. A \emph{frozen probe} trains on unedited
activations and evaluates on ablated activations. A \emph{retrained probe} trains and
evaluates on ablated activations, using the same partitions. The former measures sensitivity
of the original classifier to ablation; the latter tests whether language can still be decoded
after ablation \citep{elazar2021amnesic}. Appendix~\ref{app:probe-details} gives the fitting details.

Figure~\ref{fig:probe-depth} shows that the accuracy of the retrained probe remains above 99.7\% across the
tested depth, rank, and strength settings, while the accuracy of the frozen probe varies substantially. After
ablation of the language components, successful classification uses the component outside the removed span. Frozen-probe
sensitivity also differs across models: at the designated \MATHD{} capture minima, rank-10
\AlgOne{} projection reduces accuracy to 24.3--35.7\% for Qwen2.5, Qwen3, and GLM, whereas
R1-Distill retains 94.6\% (Table~\ref{tab:probe-supplement}).

\Pooled{} also reduces frozen accuracy substantially for the first three models, while the
rank-matched random control retains 100\% accuracy at the designated layers. Disrupting a
frozen classifier therefore does not distinguish the language-based subspaces from directions
fitted without language labels. Across depth, frozen \AlgOne{} accuracy is lowest at layer 0
or 1, rather than at the intermediate-layer capture minima.

Language identity thus remains linearly decodable after full ablation of either estimated
subspace. These local measurements do not establish whether the surviving information contributes
to generation. We examine generation outcomes next.

\section{Influencing generation through language subspaces}
\label{sec:results}
\label{sec:intervention}

We now turn from representation to behavior: what changes when these subspaces are edited during
generation? We map the intervention design space, reassess earlier results under a stricter task
criterion, and compare configurations by answer accuracy, response-language fidelity, and joint
success. Appendix~\ref{app:reference} gives the complete intervention and generation protocol.

\begin{figure}[t]
  \centering
  \includegraphics[width=\columnwidth]{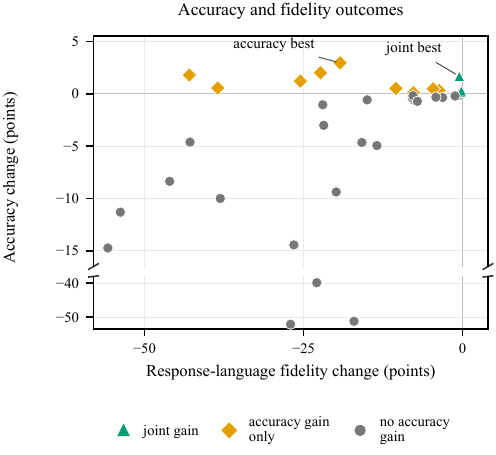}
  \caption{Accuracy gains usually lie far to the left: the intervention appears to reason better
  precisely when it abandons the requested language. Only a narrow corner improves the joint
  objective.}
  \label{fig:grid}
\end{figure}

\begin{table}[t]
  \centering
  {\small\begin{tabular}{@{}lrrr@{}}
\toprule
\multicolumn{1}{l}{Setting} & \multicolumn{1}{c}{Accuracy} &
  \multicolumn{1}{c}{Fidelity} & \multicolumn{1}{c}{Joint} \\
& \multicolumn{1}{c}{(\%) $\uparrow$} & \multicolumn{1}{c}{(\%) $\uparrow$} &
  \multicolumn{1}{c}{(\%) $\uparrow$} \\
\midrule
No intervention & 62.80 & 99.82 & 62.62 \\
\addlinespace[3pt]
\shortstack[l]{\AlgOne{}, rank 10\\projection, dose 0.5} & 65.75 & 80.58 & 53.82 \\
\addlinespace[3pt]
\shortstack[l]{\AlgOne{}, rank 1\\projection, dose 1} & 64.47 & 99.31 & 64.22 \\
\bottomrule
\end{tabular}
}
  \caption{No intervention and the \AlgOne{} configurations with the highest accuracy and joint
  accuracy on Qwen2.5 across eleven languages.}
  \label{tab:grid}
\end{table}

\begin{table*}[t]
  \centering
  {\setlength{\tabcolsep}{3pt}\small\begin{tabular}{@{}lrrrrrr@{}}
\toprule
& \multicolumn{2}{c}{Causal intervention\textsuperscript{*}} &
  \multicolumn{2}{c}{English switch} & \multicolumn{2}{c}{Chinese switch} \\
\cmidrule(lr){2-3}\cmidrule(lr){4-5}\cmidrule(l){6-7}
Model & Accuracy & Fidelity & Accuracy & Fidelity & Accuracy & Fidelity \\
\midrule
Qwen2.5 & $+0.94$ & $+4.69$ & $+1.78$ & $-64.71$ & $+3.42$ & $-95.45$ \\
Qwen3 (thinking) & $+1.27$ & $+0.03$ & $+1.87$ & $-80.44$ & $-1.73$ & $-77.33$ \\
Qwen3 (no thinking) & -- & -- & $+1.42$ & $-72.36$ & $+0.36$ & $-50.89$ \\
R1-Distill & $+2.40$ & $+1.20$ & $+1.07$ & $-22.49$ & $+2.58$ & $-31.09$ \\
GLM-Z1 & $+1.52$ & $+0.08$ & $+6.53$ & $-76.76$ & $-0.27$ & $-72.76$ \\
\bottomrule
\end{tabular}
}
  \caption{Accuracy and fidelity changes under causal intervention and prompt-language switches.
  \textsuperscript{*}Causal-intervention values are imported from \citet{zhao2025less}; switch
  values average the nine non-English/Chinese \MGSM{} languages. Comparable accuracy gains thus
  conceal radically different language behavior.}
  \label{tab:prompt}
\end{table*}

\subsection{Defining the intervention design space}
We construct the intervention in stages. First, we estimate a language-specific subspace using
either \AlgOne{} or \DiffPCA{}. For each construction, we vary the rank to test how large the
subspace must be: a low-rank edit targets a compact language-related direction, whereas a
higher-rank edit removes a broader span of activation variation.

Given a subspace, we then vary how it is edited during generation. Qwen2.5 has 28 transformer
layers, numbered 0--27. Following \citet{zhao2025less}, layers 10--19 form the middle band and
layers 20--27 the upper band. This split treats mid-network language variation as interference to
reasoning, while reserving the upper layers for expressing the result in a particular language. At
each edited layer, we apply
\begin{equation}
    h' = h - \lambda\,\delta(h),
    \label{eq:generation-edit}
\end{equation}
where $\delta$ is the intervention operator and $\lambda$ its signed dose. The reference schedule
uses $\lambda=0.3$ in the middle band and $\lambda=-0.3$ in the upper band, so the selected language
component is diminished in the middle layers and amplified in the upper layers. Orthogonal projection sets $\delta(h)=UU^\top h$; \LEACE{}
motivates covariance-aware oblique removal with minimal distortion under its guardedness objective
\citep{belrose2023leace}, while our \Wiener{}-style operator shrinks directions according to their
estimated language signal-to-noise ratio. We vary the operator and dose, calibrating dose on
\MATHD{} activations without observing \MGSM{} outcomes. This design lets us ask whether an effect
is robust to how the subspace is defined,
how much of it is retained, and how aggressively it is edited, rather than being tied to one
particular intervention recipe. Appendix~\ref{app:operators} defines the operators and complete
grid.

\subsection{Interpreting prior intervention results}
On Qwen2.5, \citet{zhao2025less} report gains in both mean \MGSM{} accuracy and response-language
fidelity using language-wise coefficient tuning under a composite objective
(Appendix~\ref{app:reference}). We instead evaluate fixed configuration procedures by whether
correctness and target-language output improve on the same responses; because the protocols
differ, our fidelity losses do not reinterpret their published gains.

\subsection{Accuracy gains can mask language substitution}
Figure~\ref{fig:grid} reveals that accuracy gains concentrate where response-language fidelity
falls. The accuracy-maximizing configuration exemplifies this trade-off (Table~\ref{tab:grid}).
Accuracy alone therefore cannot tell whether an intervention makes the model more capable or
merely changes the language in which that capability is expressed. To diagnose the difference,
we define \emph{joint accuracy}: a response succeeds only when its answer is correct and it
remains in the original problem language.

Viewed through this joint criterion, the accuracy-maximizing intervention loses its apparent
practical advantage: its gain in numerical correctness is accompanied by enough language drift
that it produces fewer answers satisfying the task as posed. The intervention has not simply
improved behavior; it has traded one part of the task contract against another.

The reversal is most striking for the simplest configuration. A rank-one \AlgOne{} subspace
with ordinary orthogonal projection yields a modest joint gain, whereas broader or more
elaborate edits do not. This cautions against equating stronger erasure with a cleaner causal
intervention. A narrow edit may perturb a behaviorally relevant direction while preserving the
distributed features that route generation into the requested language; a broader removal may
cut across features jointly involved in reasoning and language control. The grid does not by
itself identify that mechanism, but it points to \emph{selectivity}, rather than intervention
strength, as the quantity that matters. Appendix Figure~\ref{fig:grid-heatmap} shows the joint
criterion across all 37 configurations.

\section{Prompt language as a behavioral control}
\label{sec:prompt}

Language-subspace ablation is motivated by a mechanistic hypothesis: language-specific
variation in middle-layer activations may behave as noise relative to a shared reasoning
computation. If that shared computation is disproportionately shaped by high-resource
languages, removing the source-language component could move lower-resource activations toward
a regime the model uses more effectively in English or Chinese. The same movement, however,
has an obvious failure mode: it may pull the language of the output toward that high-resource
anchor as well.

We test this possibility without editing activations. In a \emph{prompt-language switch}, the
math problem remains in its original language, but its worked examples and answer cue are
replaced with English or Chinese versions. This asks whether placing the model in a
high-resource linguistic context is sufficient to reproduce both sides of the intervention:
greater access to reasoning behavior and drift away from the requested output language.
Appendix~\ref{app:prompt} illustrates the exact prompt transformation and gives the generation
protocol.

Table~\ref{tab:prompt} places these prompt-language switches beside the gains reported for the
best causal interventions of \citet{zhao2025less}. The comparison is diagnostic rather than a
leaderboard because the generation protocols differ. Its significance is that an accuracy gain
of comparable scale can be produced by contextual language routing alone. Yet the resulting
answers follow the switched prompt language, so accuracy rises while joint accuracy falls. A
better numerical score therefore does not, by itself, identify a better reasoning mechanism; it
may instead reveal that the model has entered a linguistic regime in which it is easier to
answer but less able to satisfy the original task.

This raises a simple question: what if we explicitly ask the model to answer in the target
language? We test this on GLM-Z1 while retaining the English prompt-language switch. Relative to
the native-prompt reference, whose accuracy and fidelity are 70.76\% and 94.53\%, the explicit
instruction yields 74.62\% accuracy and 97.02\% fidelity. The result suggests that the linguistic
context used to elicit computation and the language used to realize the answer can be partially
decoupled at inference time. We do not optimize the instruction further, since that would turn the
analysis into prompt engineering; here, it serves as evidence that language substitution is not
inevitable.

\section{Conclusion}
Estimated language subspaces are easy to overinterpret. Subspaces built from multilingual means or paired translation differences recover translation-associated structure, yet projection leaves language linearly decodable, and geometric salience does not predict behavioral leverage. A subspace can therefore be a useful coordinate system without being a container for the feature, or the mechanism controlling it.

The behavioral results expose the consequence. The edit that appears to improve reasoning most also abandons the requested language; prompt switches reproduce this pattern, and an explicit instruction can reverse it. Apparent gains may thus reflect entry into an easier linguistic regime rather than improved language-independent computation. Multilingual interventions must be judged against the whole task contract. The goal is not to make models reason in one privileged language, but to separate access to useful computation from the language in which users receive it.
\label{end:body}

\section*{Limitations}

Our conclusions are bounded by task and model scope. \MATHD{} and \MGSM{} are both mathematical
reasoning benchmarks, and all four models are similarly scaled at 7--9 billion
parameters. Whether the findings generalize across domains or to larger models remains open.

Additionally, the rank-one intervention was selected and evaluated on the same benchmark. Its
apparent joint advantage therefore requires held-out confirmation.

\section*{Ethical considerations}

Our analysis of Zhao et al.'s coefficient-selection procedure is based on their released analyzer.
Because the release does not identify the exact code revision used for the paper, we treat this as
a finding about the public implementation---not a claim about how the published numbers were
produced.

\section*{Declaration on Generative AI}

Generative AI systems, including Claude and OpenAI Codex, assisted with manuscript editing and
verification. They did not conduct the experiments; the authors remain responsible for the work.

\bibliography{references}

\clearpage
\newpage
\appendix

\section{Subspace and operator details}
\label{app:construction}

\subsection{Subspace construction}

\begin{table*}[th]
\centering
\begingroup
\small
\raggedright
\setlength{\parindent}{0pt}
\setlength{\parskip}{0pt}
\renewcommand{\arraystretch}{1.05}
\hrule height 0.6pt\vspace{3pt}
\textbf{In:} At fixed layer $k$, activations $h_{i,k}^{\ell}$ for $n$ problems and $L$ languages;
rank $r$ ($r\leq L-1$ for (a)). Method (b) requires parallel activations.
\par\vspace{4pt}\hrule height 0.3pt\vspace{5pt}
\noindent\begin{minipage}[t]{0.48\linewidth}\raggedright\vspace{0pt}
\textbf{(a) \AlgOne{}:}\\[1pt] Common-specific decomposition
\end{minipage}\hfill
\begin{minipage}[t]{0.48\linewidth}\raggedright\vspace{0pt}
\textbf{(b) \DiffPCA{}:}\\[1pt] Within-problem PCA
\end{minipage}\par\vspace{5pt}
\noindent\begin{minipage}[t]{0.48\linewidth}\raggedright\vspace{0pt}
\begin{tabularx}{\linewidth}{@{}>{\color{black!65}\footnotesize}r@{\hspace{0.65em}}>{\raggedright\arraybackslash}X@{}}
& {\footnotesize\itshape Approximate the language means at rank $r$.}\\[2pt]
1 & $m_\ell\leftarrow \frac{1}{n}\sum_i h_{i,k}^\ell$;\quad $W\leftarrow[m_1,\ldots,m_L]$.\\[1.5pt]
2 & $a_0\leftarrow \frac{1}{L}W\mathbf{1}$;\quad $U\Sigma V^\top\leftarrow\operatorname{SVD}(W-a_0\mathbf{1}^\top)$.\\[1.5pt]
3 & $\widehat W\leftarrow a_0\mathbf{1}^\top+U_r\Sigma_rV_r^\top$.\\[1.5pt]
\end{tabularx}
\end{minipage}\hfill
\begin{minipage}[t]{0.48\linewidth}\raggedright\vspace{0pt}
\begin{tabularx}{\linewidth}{@{}>{\color{black!65}\footnotesize}r@{\hspace{0.65em}}>{\raggedright\arraybackslash}X@{}}
& {\footnotesize\itshape Remove each problem's mean and normalize its translation residuals.}\\[2pt]
1 & $\bar h_i\leftarrow \frac{1}{L}\sum_\ell h_{i,k}^\ell$.\\[1.5pt]
2 & $R_i^\ell\leftarrow h_{i,k}^\ell-\bar h_i$;\quad $z_i^\ell\leftarrow \dfrac{R_i^\ell}{\max(\lVert R_i^\ell\rVert,\epsilon)}$.\\[1.5pt]
\end{tabularx}
\end{minipage}\par\vspace{5pt}\noindent\begin{minipage}[t]{0.48\linewidth}\raggedright\vspace{0pt}
\begin{tabularx}{\linewidth}{@{}>{\color{black!65}\footnotesize}r@{\hspace{0.65em}}>{\raggedright\arraybackslash}X@{}}
& {\footnotesize\itshape Refine the shared anchor and recover the language basis.}\\[2pt]
4 & $a\leftarrow(\widehat W^+)^\top\mathbf{1}$;\quad $a\leftarrow \dfrac{a}{a^\top a}$.\\[3pt]
5 & $\widetilde U\widetilde\Sigma\widetilde V^\top\leftarrow\operatorname{SVD}(\widehat W-a\mathbf{1}^\top)$.\\[1.5pt]
6 & \textbf{Return} the first $r$ columns of $\widetilde U$.\\[1.5pt]
\end{tabularx}
\end{minipage}\hfill
\begin{minipage}[t]{0.48\linewidth}\raggedright\vspace{0pt}
\begin{tabularx}{\linewidth}{@{}>{\color{black!65}\footnotesize}r@{\hspace{0.65em}}>{\raggedright\arraybackslash}X@{}}
& {\footnotesize\itshape Fit centered PCA to the normalized residuals.}\\[2pt]
3 & Stack the $z_i^\ell$ as rows of $Z$;\quad $Z_c\leftarrow Z-\mathbf{1}\bar z^\top$.\\[1.5pt]
4 & $A\Sigma V^\top\leftarrow\operatorname{SVD}(Z_c)$.\\[1.5pt]
5 & \textbf{Return} the first $r$ columns of $V$.\\[1.5pt]
\end{tabularx}
\end{minipage}\par\vspace{4pt}\hrule height 0.3pt\vspace{4pt}\noindent\begin{minipage}[t]{0.48\linewidth}\raggedright\vspace{0pt}
\textbf{Out:} Orthonormal language basis $\widetilde U_r\in\mathbb{R}^{d\times r}$.
\end{minipage}\hfill
\begin{minipage}[t]{0.48\linewidth}\raggedright\vspace{0pt}
\textbf{Out:} Orthonormal translation-residual basis $V_r\in\mathbb{R}^{d\times r}$.
\end{minipage}\par\par\vspace{4pt}\hrule height 0.6pt\vspace{5pt}\endgroup
\caption{Layerwise construction of \AlgOne{} and \DiffPCA{} subspaces. Subscript $r$ selects
the leading $r$ singular components; $W^+$ is the Moore--Penrose pseudoinverse and
$\epsilon>0$ is a numerical norm floor.}
\label{tab:subspace-algorithms}
\end{table*}

As Table~\ref{tab:subspace-algorithms} makes explicit, the \AlgOne{} output is the orthonormal
language basis $\widetilde U_r$, not the shared anchor
$a$. Our implementation matches the released implementation we inspected for
\citet{zhao2025less}. \DiffPCA{} is
an angular construction because each problem--language residual has equal norm before centering;
this prevents a few unusually large activations from dominating the fit. \Pooled{} applies the
same normalized PCA to all activation rows without residualizing problem identity or observing
language labels. \Rand{} is obtained by QR decomposition of a Gaussian matrix with a fixed seed.
All bases are fitted independently at each layer.

\subsection{Edit operators}
\label{app:operators}

All generation edits have the affine form $h' = h-\lambda\delta(h)$. Orthogonal projection
uses $\delta(h)=UU^\top h$. Our \LEACE{}-style operator estimates the pooled covariance
$\Sigma$, shrinks it by 0.1 toward a scaled identity, and orthonormalizes the target span after
whitening. Writing $\mu$ for the pooled mean and $Q$ for the resulting orthonormal basis, it
applies
\begin{equation}
 \delta(h)=\Sigma^{1/2}QQ^\top\Sigma^{-1/2}(h-\mu).
\end{equation}
With the centered language-mean span and full rank, this is the minimal-MSE linear guardedness
map of \citet{belrose2023leace}. With a supplied \DiffPCA{} basis, it is an oblique removal of
that span and does not inherit the supervised guardedness guarantee.

The \Wiener{} operator uses the random-effects decomposition
$h=\mu+c_i+g_\ell+q_{i\ell}+\epsilon$, where $c_i$, $g_\ell$, $q_{i\ell}$, and $\epsilon$
denote problem, language, problem--language interaction, and residual effects. For the
mean-based target, its edit is
\begin{equation}
 \delta(h)=\Sigma_g(\Sigma_g+\Sigma_c+\Sigma_\epsilon)^{-1}(h-\mu),
\end{equation}
with the covariance terms estimated from the parallel \MATHD{} activations. This attenuates
directions according to their estimated signal-to-noise ratio instead of assigning every
direction the same coefficient. For the \DiffPCA{} cross, the basis directions are scaled by
their residual-covariance energy before the same marginal-covariance shrinkage. Mean-based
\Wiener{} is intrinsically rank at most $L-1$ and has no separate rank knob. Consequently the
cross is deliberately unbalanced: the 37 cells comprise 6 mean-projection, 6 mean-oblique,
4 mean-Wiener, and 7 cells for each \DiffPCA{} operator. It should not be read as a complete
factorial comparison of estimator and operator.

\section{Representation analyses: protocols and supplementary results}
\label{app:geometry}

\begin{table}[!ht]
  \centering
  {\setlength{\tabcolsep}{2pt}\small\begin{tabular}{@{}lrrrr@{}}
\toprule
Model/data & \AlgOne{} & \DiffPCA{} & $k$/depth & $r_{\mathrm{P}}$ \\
\midrule
Qwen2.5/\MATHD{} & 0.431 & 0.479 & 18/0.67 & 0.9971 \\
Qwen2.5/\MGSM{} & 0.297 & 0.352 & 18/0.67 & 0.9918 \\
Qwen3/\MATHD{} & 0.592 & 0.608 & 21/0.60 & 0.9995 \\
Qwen3/\MGSM{} & 0.448 & 0.462 & 21/0.60 & 0.9972 \\
R1/\MATHD{} & 0.459 & 0.494 & 16/0.59 & 0.9940 \\
R1/\MGSM{} & 0.240 & 0.296 & 17/0.63 & 0.9882 \\
GLM-Z1/\MATHD{} & 0.415 & 0.479 & 20/0.51 & 0.9976 \\
GLM-Z1/\MGSM{} & 0.328 & 0.395 & 20/0.51 & 0.9911 \\
\bottomrule
\end{tabular}
}
  \caption{Rank-10 translation capture at each model--corpus pair's \AlgOne{} interior
  minimum. $k$ is the block index; depth is $k/(K-1)$. $r_{\mathrm{P}}$ is the Pearson
  correlation between the two estimators' full depth profiles.}
  \label{tab:trough}
\end{table}

\begin{table*}[th]
  \centering
  {\small\begin{tabular}{@{}lrccccrr@{}}
\toprule
Model & $k$ & \AlgOne{} & \DiffPCA{} & \Pooled{} & \Rand{} & Perm. & Min. retr. \\
\midrule
Qwen2.5 & 18 & 0.357 / 1.000 & 0.441 / 1.000 & 0.364 / 1.000 & 1.000 / 1.000 & 0.086 & 0.9993 \\
Qwen3 & 21 & 0.243 / 1.000 & 0.296 / 1.000 & 0.186 / 1.000 & 1.000 / 1.000 & 0.092 & 0.9971 \\
R1-Distill & 16 & 0.946 / 1.000 & 0.925 / 1.000 & 0.992 / 1.000 & 1.000 / 1.000 & 0.089 & 0.9985 \\
GLM-Z1 & 20 & 0.337 / 1.000 & 0.494 / 1.000 & 0.291 / 1.000 & 1.000 / 1.000 & 0.093 & 0.9993 \\
\bottomrule
\end{tabular}
}
  \caption{Language probes at the held-out \MATHD{} capture minimum, using rank ten and
  full projection ($\lambda=1$). Each method cell gives frozen / retrained held-out accuracy.
  Clean accuracy is 1.000 for every model. The permutation reference uses shuffled labels on clean
  activations. The final column is the lowest retrained accuracy over all tested intervention
  cells across the depth, rank, and strength sweeps, reported to four decimals.}
  \label{tab:probe-supplement}
\end{table*}

\begin{figure*}[t]
  \centering
  \includegraphics[width=\textwidth]{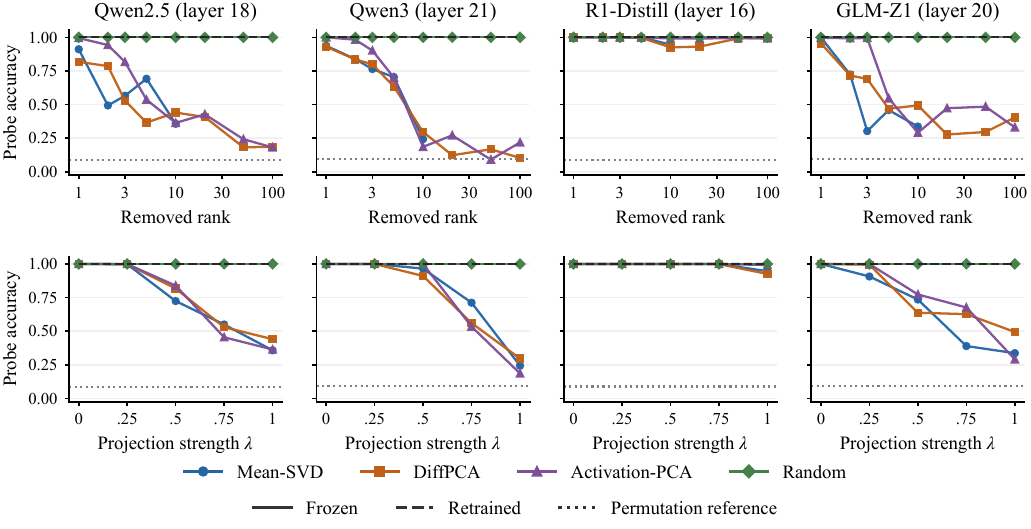}
  \caption{Probe sensitivity at each model's held-out \MATHD{} capture minimum. Top:
  rank varies at full projection. Bottom: strength varies at rank ten. Solid curves are frozen
  probes; dashed curves are retrained probes and overlap near 1. The dotted gray line is the
  measured permutation reference at that layer. \AlgOne{} ends at rank ten.}
  \label{fig:probe-sweeps}
\end{figure*}

\subsection{Activation extraction and data partitions}

Representation analyses use zero-shot residual-stream activations at the final prompt token;
the eight demonstrations used in generation are absent. Layers are indexed from zero over
transformer block outputs, with relative depth $k/(K-1)$ for a model with $K$ blocks.
Qwen3 uses non-thinking mode. GLM's chat template appends an opening thinking prefix, so its
final prompt token follows that prefix. Section~\ref{app:boundary} examines this choice.

For held-out \MATHD{} capture, we split the 7,500 problems into 3,750 fitting and 3,750
evaluation problems using seed zero. All translations of each problem stay in the same
partition. For cross-corpus capture, we fit on all 7,500 \MATHD{} problems and evaluate on
250 \MGSM{} problems. The fitting sample size and evaluation corpus both change between these
tests, so their absolute capture levels do not isolate a corpus effect.

The probe experiment uses a third partitioning scheme: subspaces are fitted on all 7,500
\MATHD{} problems, but classifiers are trained and evaluated within \MGSM{} using grouped
cross-validation. Thus the subspace transfers across corpora; the classifier itself is not
trained on \MATHD{} and transferred to \MGSM{}.

\subsection{Translation-capture results}

Capture is the mean of the per-vector squared-norm fractions in Equation~\ref{eq:capture},
with equal weight for every evaluation problem and each of the 55 unordered language pairs.
It is not a ratio of pooled squared norms. Table~\ref{tab:trough} reports both estimators at
the layer minimizing rank-10 \AlgOne{} capture within $0.15\leq k/(K-1)\leq0.95$, selected
separately for each model and evaluation corpus. Only Qwen3/\MATHD{} differs from the
unrestricted minimum, which occurs at layer 1. The correlations summarize agreement between
full depth profiles, not alignment of the underlying subspace directions.

\paragraph{Rank sensitivity.}
At Qwen2.5 layer 18 on held-out \MATHD{}, ranks 10, 20, and 100 yield \DiffPCA{}
capture of 0.479, 0.565, and 0.740, versus 0.0028, 0.0056, and 0.0276 for random spans.

\subsection{Language-probe protocol}
\label{app:probe-details}

We classify eleven languages using logistic regression on unscaled activations. The
implementation calls scikit-learn's \texttt{LogisticRegression} with $C=1.0$,
\texttt{max\_iter=3000}, and \texttt{random\_state=0}; other arguments use library defaults.
Five-fold \texttt{GroupKFold} partitions the 250 \MGSM{} problem indices. Each fold trains
on 200 problems (2,200 activations) and evaluates on 50 (550 activations), keeping all eleven
translations together. Accuracy is averaged over the five held-out folds.

The clean probe trains and evaluates on unedited activations. Within the same folds, the frozen
probe trains on clean training activations and evaluates on projected held-out activations;
the retrained probe trains and evaluates on projected activations. Every edit is applied
analytically at the measured layer using Equation~\ref{eq:ablation}. No edited state is
propagated through subsequent blocks. Subspace fitting never uses the held-out probe examples.

The three sweeps vary one factor at a time:
\begin{itemize}
  \item \textbf{Depth:} every block, with rank ten and $\lambda=1$, for \AlgOne{} and \DiffPCA{}.
  \item \textbf{Rank:} $r\in\{1,2,3,5,10,20,50,100\}$ at the designated layer and
  $\lambda=1$, for both estimators and the \Pooled{} and \Rand{} controls. \AlgOne{} is
  restricted to $r\leq10$.
  \item \textbf{Strength:} $\lambda\in\{0,0.25,0.5,0.75,1\}$ at the designated layer
  and rank ten, for both estimators and both controls.
\end{itemize}
The designated layers are the rank-10 \AlgOne{} interior minima on held-out \MATHD{}:
18, 21, 16, and 20 for Qwen2.5, Qwen3, R1-Distill, and GLM, respectively.
In particular, R1-Distill's probe sweeps use layer 16, even though its \MGSM{} capture
minimum is at layer 17. The sweeps do not cover the full depth--rank--strength product.

For the permutation reference, we globally shuffle the language-label vector five times
using seed zero and repeat grouped cross-validation on clean activations. We measure this
reference at the first, designated, and final layers. In Figure~\ref{fig:probe-depth},
unmeasured layers use the mean of these three measured values; the gray reference is therefore
not independently measured at every layer. The references in Table~\ref{tab:probe-supplement}
and Figure~\ref{fig:probe-sweeps} are measured at the designated layer. They provide an
empirical chance reference, not a statistical lower bound on accuracy.

\subsection{Probe results and controls}

Table~\ref{tab:probe-supplement} reports the model-level results. The rank sweeps in
Figure~\ref{fig:probe-sweeps} show that frozen accuracy need not fall monotonically with
the number of removed dimensions: R1-Distill's frozen \DiffPCA{} probe has accuracy 0.925
at rank ten and 0.995 at rank 100. Across all tested intervention cells, retrained accuracy
remains at least 0.9971, and the random control retains 1.000 accuracy in every tested rank
and strength condition.

\subsection{GLM extraction-boundary sensitivity}
\label{app:boundary}

GLM's forced-thinking template ends with an opening \texttt{<think>} prefix. On 64
\MGSM{} problems per language, we compare block outputs at the token immediately before
this prefix with those at the final token after it, within the same forward pass. For
representations $a$ and $b$ at these positions, the diagnostics are cosine similarity and
relative change $\lVert b-a\rVert/\lVert a\rVert$. We first average over problems within
each language and layer, then equally over all eleven languages and 40 layers.

The mean cosine similarity is 0.645 and the mean relative change is 0.865. Per-language
cosine similarities, averaged over layers, range from 0.634 to 0.661. Capture and probe
analyses use the post-prefix position; their robustness across the two extraction positions
was not tested.

\section{Intervention selection and evaluation}
\label{app:reference}

The reference schedule uses the middle/high layer bands of \citet{zhao2025less} and one fixed
point within their coefficient search grid: $+0.3$ at layers 10--19 and $-0.3$ at layers 20--27.
Their Appendix~F specifies the coefficient grid for Qwen2.5-7B. The released analyzer, pinned to
the inspected revision in \citet{zhao2025code}, takes the no-intervention cell as the baseline and,
among intervention cells, selects a separate pair for each language by maximizing
\begin{equation}
 2(A-A_0)+(F-F_0),
\end{equation}
where all scores are proportions and $A_0$ and $F_0$ are the no-intervention scores. Selection
excludes the baseline and is performed independently for each language.
Thus ``reference'' in our grid means our single fixed $+0.3/-0.3$ schedule applied to every
language, not a coefficient pair reported as universal by Zhao et al. Off-center schedules are
calibrated using activation-edit energy on \MATHD{}, without observing \MGSM{} outcomes.

For each construction--operator pair, dose 1 is the reference energy. The other dose rungs
target $0.5$, $1.5$, and $2$ times that energy by changing the coefficient magnitude for each
language. Both layer bands are scaled together. Rank rungs are evaluated at the central
coefficients; dose rungs use rank ten. \DiffPCA{} additionally admits rank 32, whereas
\AlgOne{} is capped at ten. All intervention generations use a 512-token limit. Edits operate on
every prompt token during prefill and are removed before autoregressive decoding begins.

\begin{figure*}[t]
  \centering
  \includegraphics[width=\textwidth]{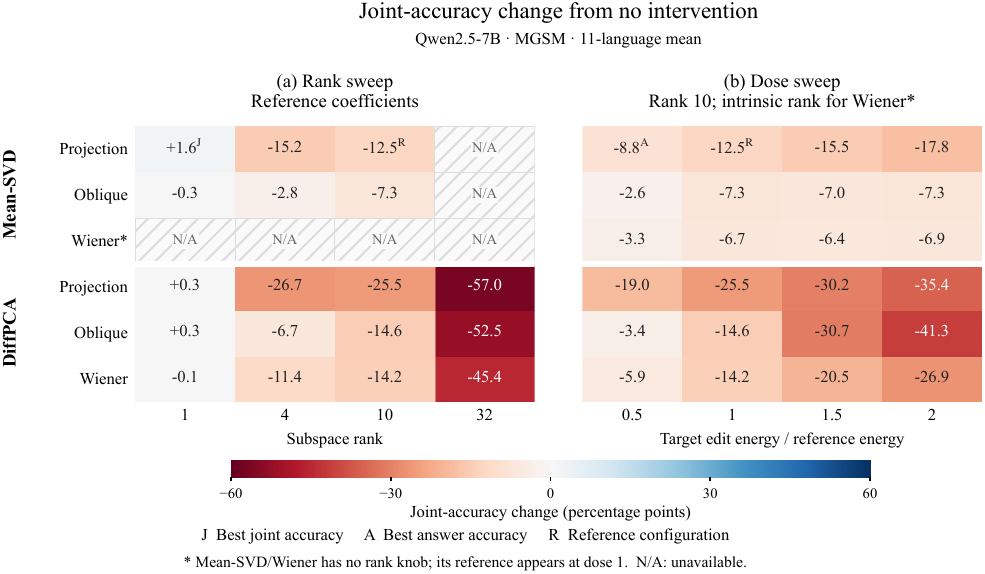}
  \caption{Joint-accuracy changes across 37 intervention configurations. Both panels
  use the same symmetric color scale. Five reference settings appear in both sweeps;
  Mean-SVD/Wiener's intrinsic-rank reference appears only at dose 1. Letters identify the
  best joint accuracy (J), best answer accuracy (A), and
  reference projection (R).}
  \label{fig:grid-heatmap}
\end{figure*}

\subsection{Scoring and uncertainty}
\label{app:stats}

Answers are extracted after any completed visible reasoning block. Native decimal digits are
mapped to ASCII, and the last localized answer-field number is compared numerically with the
gold answer. To classify response language, we first strip mathematics. Kana identifies
Japanese, while dominant Han, Cyrillic, Bengali, Telugu, or Thai script identifies the
corresponding language. Remaining Latin prose must contain at least eight alphabetic characters;
we then take the highest-ranked \texttt{langdetect} candidate among German, English, Spanish,
French, and Swahili, with \texttt{DetectorFactory.seed=0}. A missing or unidentifiable label
receives $F_i=0$ in the strict metric.

Each pooled statistic first averages problem indices within a language and then averages
languages equally. For intervals and tests, 4,000 paired bootstrap replicates resample the 250
matched problem indices; every draw carries together all languages and all conditions being
contrasted for an index. Reported 95\% intervals are pointwise percentile intervals, not
simultaneous intervals over the grid. Approximate two-sided bootstrap $p$-values compare each
intervention with baseline using twice the smaller proportion of nonpositive or nonnegative
bootstrap differences, capped at one.
For the reported all-language comparisons, Holm adjustment uses one family of 37 intervention
arms for each of accuracy, fidelity, and joint accuracy. We flag a cell when
its truncation rate differs from baseline by more than 0.05; this is a sensitivity gate, not a
correction for high truncation shared with the baseline.

\subsection{Grid-wide outcomes and truncation sensitivity}

Across the 407 configuration--language cells, 152 have positive accuracy changes and 110 have
positive joint changes. After averaging all eleven languages, 12 of 37 configurations exceed
baseline accuracy and three exceed baseline joint accuracy. One comparison for each metric survives
Holm correction. Selection and evaluation use the same benchmark; no independent confirmation
set is used.

Baseline Telugu and Bengali responses reach the 512-token limit in 78.8\% and 36.8\% of cases,
respectively. Of the 407 cells, 112 change truncation by more than five percentage points.

\begin{table*}[th]
  \centering
  {\small\setlength{\tabcolsep}{5pt}\begin{tabular}{@{}lrrrrrr@{}}
\toprule
& \multicolumn{3}{c}{Prompt-language switch} & \multicolumn{3}{c}{Translate-test} \\
\cmidrule(lr){2-4}\cmidrule(l){5-7}
Model & Accuracy & Fidelity & Joint & Accuracy & Fidelity & Joint \\
\midrule
\multicolumn{7}{@{}l}{\textbf{English}} \\
\addlinespace[2pt]
Qwen2.5 & $+1.78$ & $-64.71$ & $-40.49$ & $+27.02$ & $-99.82$ & $-61.33$ \\
Qwen3 (thinking) & $+1.87$ & $-80.44$ & $-65.24$ & $+14.09$ & $-99.60$ & $-82.04$ \\
Qwen3 (non-thinking) & $+1.42$ & $-72.36$ & $-53.29$ & $+20.04$ & $-99.96$ & $-74.89$ \\
R1-Distill & $+1.07$ & $-22.49$ & $-6.71$ & $+32.58$ & $-97.85$ & $-48.71$ \\
GLM-Z1 & $+6.53$ & $-76.76$ & $-55.73$ & $+25.91$ & $-94.49$ & $-70.40$ \\
\midrule
\multicolumn{7}{@{}l}{\textbf{Chinese}} \\
\addlinespace[2pt]
Qwen2.5 & $+3.42$ & $-95.45$ & $-58.22$ & $+19.51$ & $-99.82$ & $-61.33$ \\
Qwen3 (thinking) & $-1.73$ & $-77.33$ & $-63.16$ & $+6.36$ & $-99.60$ & $-82.04$ \\
Qwen3 (non-thinking) & $+0.36$ & $-50.89$ & $-37.87$ & $+10.84$ & $-99.96$ & $-74.89$ \\
R1-Distill & $+2.58$ & $-31.09$ & $-9.47$ & $+28.53$ & $-97.85$ & $-48.71$ \\
GLM-Z1 & $-0.27$ & $-72.76$ & $-52.76$ & $+17.51$ & $-94.53$ & $-70.44$ \\
\bottomrule
\end{tabular}
}
  \caption{Prompt-language switches and translate-test on the same nine \MGSM{} languages.
  Values are percentage-point changes from the matched native-prompt baseline. Fidelity and
  joint accuracy are scored against the original problem language, including for translate-test.
  }
  \label{tab:prompt-detail}
\end{table*}

\section{Prompt-language experiments}
\label{app:prompt}

\subsection{Prompt conditions and evaluation set}

All conditions use eight worked examples. The native condition presents the examples, question,
and answer cue in the original problem language. A switch replaces the examples and cue with
English or Chinese while leaving the question in its original language. Translate-test also
translates the question into the switch language, using the benchmark's existing translations.
In the cue control, the examples remain in the switch language but the cue is restored to the
native language. The target-instruction
control retains the switched examples and cue and prepends an explicit request for the final
answer and explanation in the native language.

Table~\ref{tab:prompt-detail} uses the same nine languages as the prompt-switch columns of
main-text Table~\ref{tab:prompt}: Bengali, German, Spanish, French, Japanese, Russian, Swahili,
Telugu, and Thai. Each condition contains 250 problems per language. Excluding both English and
Chinese fixes the evaluation set across the two switches. Scores average languages equally;
each change is relative to that model and mode's native-prompt baseline on these same problems.

\subsection{Generation and scoring}

Qwen2.5 uses greedy decoding with a 1,024-token limit. Qwen3, R1-Distill, and GLM use sampling
with a 16,384-token limit; Qwen3's thinking and non-thinking settings are reported separately.
GLM uses vLLM 0.8.5.post1, temperature 0.6, and top-$p$ 0.95. These budgets differ from the
512-token intervention runs, so comparisons use changes from each protocol's own baseline.

Accuracy measures numerical correctness, fidelity measures output in the original problem
language, and joint accuracy requires both on the same response. Unidentified responses count
as fidelity and joint failures, using the scoring procedure in Appendix~\ref{app:stats}.

\subsection{Switch and translate-test results}

Table~\ref{tab:prompt-detail} extends the main comparison with joint accuracy and both
translate-test conditions. Every prompt-language switch reduces joint accuracy, including
settings that improve numerical accuracy. Translate-test increases accuracy for every model
and mode in both languages.

\subsection{Target-language instruction and answer cue}

\begin{table}[ht]
  \centering
  {\small\setlength{\tabcolsep}{4pt}\begin{tabular}{@{}lrr@{}}
\toprule
Prompt & Seed 0 & Seed 1 \\
\midrule
Native prompt & 70.44 & 73.78 \\
English switch & 14.71 & 15.47 \\
English + target instruction & 72.49 & 72.04 \\
Chinese switch & 17.69 & 17.47 \\
Chinese + target instruction & 59.69 & 57.60 \\
\bottomrule
\end{tabular}
}
  \caption{GLM joint accuracy (\%) on the same nine languages, with and without an explicit
  target-language instruction.}
  \label{tab:prompt-target}
\end{table}

The target-language instruction restores much of the lost joint accuracy in both GLM seeds
(Table~\ref{tab:prompt-target}). 

Changing only the answer cue also affects language fidelity. With English examples, accuracy
changes by $-0.44$ to $+2.36$ points and fidelity falls by 3.97--48.80 points; with Chinese
examples, the corresponding ranges are $-0.44$ to $+2.18$ and 18.57--48.22 points.

\end{document}